\documentclass[runningheads]{llncs}
\usepackage{graphicx}
\usepackage{amsmath,amssymb} 
\usepackage{color}
\usepackage[width=122mm,left=12mm,paperwidth=146mm,height=193mm,top=12mm,paperheight=217mm]{geometry}
\usepackage{algpseudocode}
\usepackage{multicol,multirow}
\usepackage{float}
\begin{document}
\pagestyle{headings}
\mainmatter

\title{Multiple Random Masking Autoencoder Ensembles for Robust Multimodal Semi-supervised Learning} 
\titlerunning{Random Masking Autoencoder Ensembles for Semi-supervised Learning}

\author{Alexandru-Raul Todoran\textsuperscript{1} \; \; \; \; Marius Leordeanu\textsuperscript{2,3} 
}
\institute{\textsuperscript{1}Aurel Vlaicu High School Orăștie, \\ \textsuperscript{2}University Politehnica of Bucharest, \\
\textsuperscript{3}Bitdefender, Romania, \\
\small{alexandruraul.todoran@gmail.com \;   leordeanu@gmail.com}
}

\maketitle

\begin{abstract}

There is an increasing number of real-world problems in computer vision and machine learning requiring to take into consideration multiple interpretation layers (modalities or views) of the world and learn how they relate to each other. For example, in the case of Earth Observations from satellite data, it is important to be able to predict one observation layer (e.g. vegetation index) from other layers (e.g. water vapor, snow cover, temperature etc), in order to best understand how the Earth System functions and also be able to reliably predict information for one layer when the data is missing (e.g. due to measurement failure or error).

\end{abstract}

\section{Introduction}
The ability to learn from multi-modal data and find consensus between multiple interpretations of the world is vital for obtaining a holistic picture that puts together in a coherent way the multiple types of observation. As we show in this paper, such a multi-modal approach could be effectively used for handling missing data as well as learning unsupervised, when the consensual output of our proposed Multiple Random Masking Auto-encoder Ensembles (MR-MAE) becomes the supervisory signal.

The problem of making predictions in a multi-layer (multi-modal) dataset allows for various approaches. However, most existing methods are task-specific and focus on maximizing the performance of a certain predetermined prediction direction -- where the type of information given at input and the type of expected output is predetermined from the start. These methods achieve high accuracy on the specific task they were trained for, but in order to learn the complex interconnections between the various data layers, one must not limit the model to a specific input-output task, but allow it to explore the multi-layered system from all directions. In other words, our learned model should be able to predict any set of layers from any other set of layers. This would make the system more robust to noise and function even when
certain layers are missing (e.g. due to measurement failure). As we show in this work, it would also allow us to learn implicit ensembles with very large number of candidates, which would further allow for efficient semi-supervised learning.

From a modeling perspective, our proposed model is related to Masked Auto-encoders (MAE) \cite{he2022masked}. Most work with Masked Auto-encoders proved the usefulness of pretraining Transformer models by masking the input and learning to fully reconstruct it, including the masked parts. Our approach is to use this idea for training and testing as well, not just as a pretraining phase. Then, at test time, the model becomes implicitly an ensemble (with different random masking patterns producing different candidates in the ensemble), with a better and more robust performance than the initial single model. This is one key difference between our MR-MAE and the recent MAE models that consider multiple modalities~\cite{mizrahi20234m,bachmann2022multimae}.  Another key difference from the aforementioned works is that our ensembles, obtained by the repeated random masking of the same input, also provide high quality pseudo-labels for effective unsupervised learning (by taking the consensual output over the multiple outputs produced for each masked piece). 

There are many problems that are by nature multi-task and multi-modal, but there is still only a small number of published papers that offer an effective semi-supervised solution. Most unsupervised learning methods consider relations between only a limited number of types of knowledge, such as depth, relative pose or semantic segmentation ~\cite{chen2019self,zhou2017unsupervised,ranjan2019competitive,bian2019unsupervised,gordon2019depth,yang2018unsupervised,tosi2020distilled,guizilini2020semantically,chen2019towards,stekovic2020casting} or modalities ~\cite{Hu_2019_CVPR,Li_2019_CVPR,zhang2017split,pan2004automatic,he2017unsupervised,zhao2018sound}. 
On the other hand, other methods that explore the relationships between many tasks do not fully address the semi-supervised learning scenario~\cite{zamir2018taskonomy,zamir2020robust,fifty2021efficiently}. 

In more recent works~\cite{marcu2023self,pirvu2023multi,leordeanu2021semi}, authors consider the semi-supervised learning case by producing automatic pseudo-labels from ensembles formed by the multiple pathways that reach a certain type of interpretation layer in a larger hyper-graph. However, different from our approach, those methods are limited to a predetermined set of input and output layers (the prediction can only go from the predetermined input layers to output layers) and the structure of the hyper-graph is fixed and built by hand. On the contrary, our model can map any set of input layers to any set of output ones, by training, when we randomly mask set of layers. Also, this training approach by repeated random masking for the same training data sample, enables the model to learn an exponential number of input-output mappings, essentially offering ensembles with a much larger number and variety than the limited, fixed number of hyper-edges (or edges) considered in previous works.

In this paper we introduce a relatively simple and effective way to fully capture the interdependence between many data layers (modalities). While it is inspired from two different directions in the literature -- unsupervised learning with graphs and hyper-graphs and learning with Masked Auto-encoders -- our approach brings them together to exploit the benefits of both. 

\textbf{Main contributions:} we contribute along several distinct dimensions:

\begin{enumerate}
\item
\textbf{Powerful Ensembles at no Extra Cost}. We show that the idea of randomly masking the data is very powerful not just for pre-training (as shown in previous works), but also during the final training and testing phase. It actually offers, for free, a very large ensemble of models (implicitly) -- in which each random masking becomes essentially a candidate in the large ensemble pool -- the number of potential candidates is very large and resulting ensemble powerful, as our experiments demonstrate.
\item
\textbf{Full Flexibility to Map any Inputs to any Outputs}. The strategy of multiple random masking over multiple layers (modalities) and locations also offers the possibility to map any set of input layers to any set of output ones. Thus, the trained model is able to overcome any case of missing data. It is able to predict any desired output layer from any set of available layers. 

\item
\textbf{Feature Importance Estimation and Selection}. The multiple random masking also provides, for free (without any additional training), the possibility to perform feature selection -- to automatically determine the importance of each feature for predicting any output feature or an entire layer. This allows us to measure the overall importance of features from any layer and any location within a layer, which could be efficiently used for feature selection.
\item
\textbf{Semi-supervised Learning}. The very large ensembles learned (large implicit pool of candidate models within a single MAE) can also provide robust pseudo-labels for the test cases where ground truth is missing, which immediately gives the possibility of unsupervised learning on such pseudo-labels. The pseudo-labels are obtained by finding the consensus output (e.g. plain average or a more sophisticated ensemble strategy) over all the randomly masked candidates that predict a given output feature.
\item
\textbf{Direct Application to Climate Studies}. We apply our robust multi-modal approach to the Earth Observation NEO Dataset from NASA, from which we consider 19 different observation layers -- a type of data that perfectly suits our scenario, since in many cases, there are also entire layers in NEO that are entirely missing for certain months.
We validate through experiments that our approach is very efficient in learning hidden connections between values from different observation layers and at very different, distant locations. Thus, our system could become a very useful tool for climate scientists to discover hidden such connections to help better understand Climate Change and the Earth System.
\end{enumerate}

\section{Our Approach}
\subsection{Training Phase}

Initially, we describe our method in a general, more abstract context, as it can be applied to any specific model (e.g. Transformer, GNN, ConvNet or MLP). Consider a dataset $X=(\mathbf{x}_0,\mathbf{x}_1,\dots,\mathbf{x}_{k-1})$, where $k$ is the number of data points (observations). Each observation $i$ is described by $n$ features $\mathbf{x}_i=(x_{i,0},x_{i,1},\dots,x_{i,{n-1}})$. The $n$ features are partitioned into $m$ layers (modalities) $N_0,N_1,\dots,N_{m-1}$, and we will use the notation $\mathbf{x}_{i,N_j}$ for the vector including only features from the layer $N_j$ for observation $i$. So, each feature, in the general sense comes from a certain interpretation layer (or modality) and a certain location (which could be both spatial and temporal) within that layer. 

For clarity, in all figures presented below, we omit the indices of observations. This is, instead of writing $(x_{i,0},x_{i,1},\dots,x_{i,{n-1}})$, we will simply use the notation $(x_0,x_1,\dots,x_{n-1})$, where $x_0, x_1, \dots x_{n-1}$ are scalars (and therefore not in bold).

During the training stage, the task is to fully learn the interconnections of among all features (from all layers and locations) by training a Masked Auto-encoder Model that is able to predict all features of an observation, when given the same, but randomly masked features as input.

\begin{figure}[H]
\centering
\includegraphics[scale=0.25]{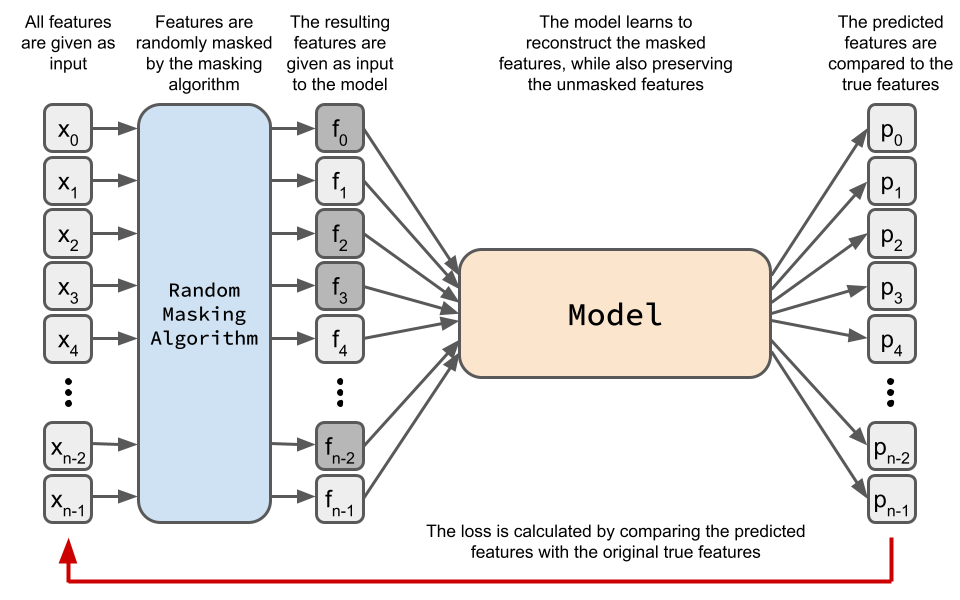}
\caption{One iteration of the learning process for training the MAE model}
\label{fig:training}
\end{figure}

The training algorithm consists of 3 steps.
\begin{enumerate}
    \item The full feature vector $\mathbf{x}=(x_0,x_1,\dots,x_{n-1})$ for a certain observation is given as input to the \textbf{Random Masking Algorithm}. The algorithm randomly chooses based on a certain probability distribution a subset $M$ of features to mask, and replaces each one of them with its mean over all observation points.
    \[
        f_j=\left\{
        \begin{array}{ll}
            x_j & j \not\in M \\
            \frac{1}{k}\sum_{i=0}^{k-1}x_{i,j} & j \in M 
        \end{array}
        \right.
    \]
    \item The resulting vector of partially masked features $\mathbf{f}=(f_0,f_1,\dots,f_{n-1})$ is then fed to the \textbf{Model}, which produces a prediction $\mathbf{p}=(p_0,p_1,\dots,p_{n-1})$.
    \item The prediction is then compared with the true unmasked features using the \textbf{Loss Function}. The loss, denoted by $Loss(\mathbf{x},\mathbf{p},M)$ is then propagated backwards using gradient descent.
\end{enumerate}

The 3 steps of the training stage include 3 components that can be subject to variations, and in our experiments, we explored different options for each of them.

1. The \textbf{Random Masking Algorithm} can follow, in general, any masking probability distribution across layers and locations. It can, for example,
mask a fixed percentage of features each time (e.g. $70\%$) or chose a different, random percentage between $0\%$ and $99\%$ each time. Additionally, masking random subsets of full layers during the learning stage is a closer representation of real-world scenarios of missing data (e.g. satellite observation failure), and can also prevent the model from relying only on data from specific layers of high relevance.

2. The \textbf{Model} is the most flexible component of our method. In this paper, we will use a Multi-Layer Perceptron (MLP), but this strategy is perfectly suited for any powerful deep learning model (e.g. modern Transformers), and testing with such powerful models would be our next step in future work. For now, to prove our points, we believe that the very general Multiple Layer Perceptrons suffices.

3. The \textbf{Loss Function} can be adjusted as well, in order to put higher emphasis on predicting the masked features, while also preserving the unmasked features.

\subsection{Automatic Feature Importance Estimation}

Next, we describe an effective and flexible way to automatically estimate the importance of each feature when making predictions. Consider the case of making predictions for an observation $i$. We repeat the process for many random masking subsets $M_0,M_1,\dots,M_{l-1}$, and for each of them, we compute $Loss(p_j,x_j)$ for each predicted feature $j$. Consider one such feature $j$, and the loss values it produced \textbf{only when it was masked} (there are a maximum of $l$ such values). The key intuition is to harness the way this loss varies over different masking subsets, and draw conclusions on which features, when \textbf{not masked}, contribute towards a lower loss for predicting another \textbf{masked} feature.

Therefore, the original method we propose consists of computing the $n$ by $n$ \textbf{Loss Matrix}. Cell $(a,b)$ contains a value representing the average loss of predicting the \textbf{masked} feature $b$, when feature $a$ is \textbf{unmasked}.
\begin{enumerate}
    \item We go through multiple iterations over all observations, large enough for the average loss values to converge.
    \item For each observation $i$ within the dataset, and its corresponding randomly generated masking subset $M$ (distinct for each iteration),
    \item and for each feature $a \not\in M$ (\textbf{unmasked}) and feature $b \in M$ (\textbf{masked}), we add $Loss(p_b,x_b)$ to cell $(a,b)$ in the \textbf{Loss Matrix}.
    \item In the end, we compute the average for each cell $(a,b)$.
\end{enumerate}

\begin{figure}[H]
\centering
\includegraphics[scale=0.3]{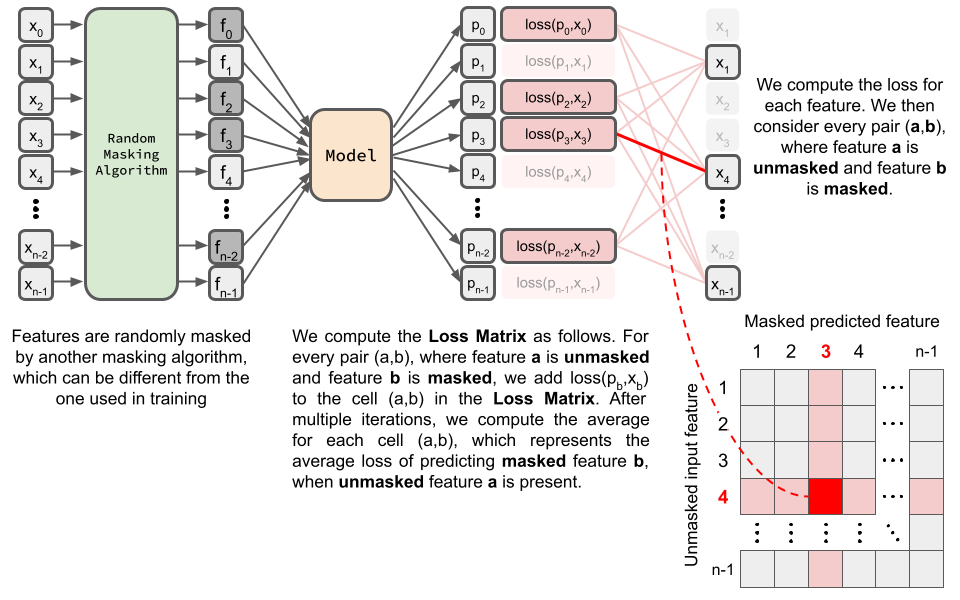}
\caption{Our method of automatically estimating feature importance by computing the {\it Loss Matrix}. The figure shows how one prediction case translates to changes in the matrix. In particular, one pair of features and its corresponding cell are highlighted}
\label{fig:loss}
\end{figure}

After the process is over, we can compute 3 types of feature importance estimations:
\begin{enumerate}
    \item \textbf{Feature-wise}. For a target feature $t$, column $t$ of the \textbf{Loss Matrix} is used to determine the importance of each feature when predicting the target feature $t$.
    \item \textbf{Layer-wise}. For a target layer $N_t$, row averages over columns $j\in N_t$ are used to determine the importance of each feature when predicting the target layer $N_t$.
    \item \textbf{Global}. The row averages of the \textbf{Loss Matrix} are used to determine the general importance of each feature when making predictions.
\end{enumerate}

Based on the global importance estimates, the opportunity for extending the algorithm to feature selection arises implicitly. Different from other ways of measuring feature importance for selection purposes, our method computes the actual, demonstrated importance when making predictions, instead of looking into the inner-workings of the model. This is why devising such a selection algorithm and comparing it with standard methods is one of our next steps in future work.

\clearpage

\subsection{Building Ensembles by Multiple Random Masking}

Now we describe a way to build ensembles over the base model in order to enhance robustness, with no extra training required. The key idea is to leverage on the outstanding adaptability of the Model. Typically, Ensemble Learning requires multiple, distinct models, learning different prediction pathways. However, using our multiple random masking approach for each data sample during training, the model already consists of a multitude of learning pathways. This property follows implicitly from the very task it was trained on: to build a reliable network of prediction pathways that can overcome missing data. In an intuitive way, two highly disjoint masks over the same input require two distinct underlying networks, which coexist within the same full Model. Therefore, due to the nature of its multiple random masking training, the Model is in fact an ensemble of nets, over all possible combinations of masked input for any given unmasked output.

\begin{figure}[H]
\centering
\includegraphics[scale=0.4]{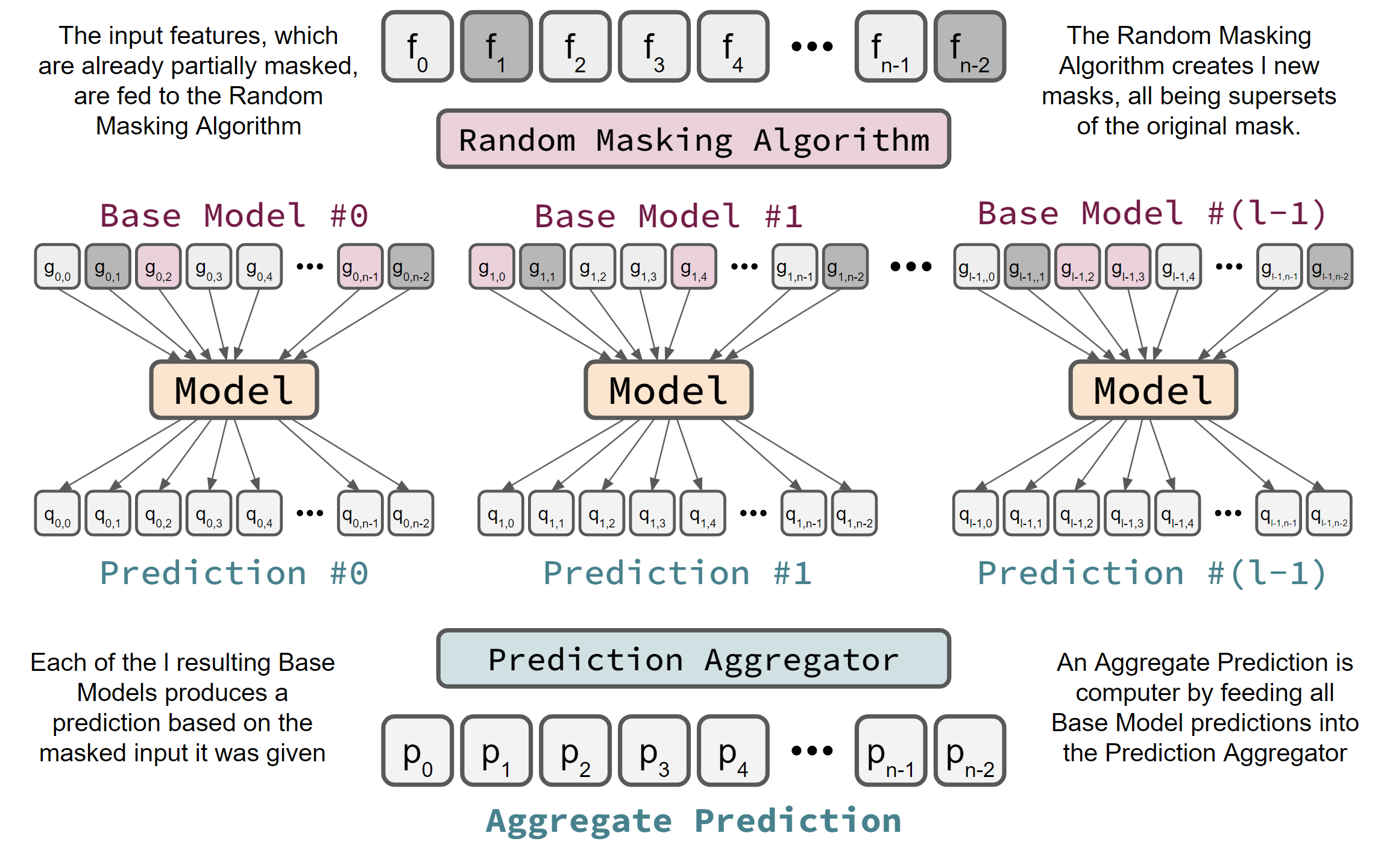}
\caption{The algorithm for constructing implicit ensembles on the base model}
\label{fig:ensemble}
\end{figure}

We therefore propose the following algorithm for computing an Aggregate Prediction for a certain input $\mathbf{x}=(x_0,x_1,\dots,x_{n-1})$. Suppose it was already masked with the mask $M$, resulting in the masked input $\mathbf{f}=(f_0,f_1,\dots,f_{n-1})$.
\begin{enumerate}
    \item We go through $l$ iterations. 
    \item For each iteration $i$, we use the \textbf{Random Masking Algorithm} to generate a random mask $M_i$, such that $M \subseteq M_i$.
    \item We use the trained \textbf{Model} to produce a prediction $\mathbf{q}_i=(q_{i,0},q_{i,1},\dots,q_{i,n-1})$.
    \item In the end, we aggregate all predictions using the \textbf{Prediction Aggregator}, thus computing the Aggregate Prediction $\mathbf{p}=Agg(\mathbf{q}_0,\mathbf{q}_1,\dots,\mathbf{q}_{l-1})$.
\end{enumerate}

Once again, this algorithm includes 2 components that can be subject to variation.
\begin{enumerate}
    \item The \textbf{Random Masking Algorithm} can be different from the one used in the training stage. Thus, choosing a fixed masking percentage or different percentages for each iteration can influence performance.
    \item The \textbf{Prediction Aggregator} can range from a simple average, to Linear Regression or even a small Neural Network. In this paper, we will use the average aggregator for simplicity, but our next steps involve experimenting with other aggregators as well.
\end{enumerate}

\section{Experiments on NASA's Earth Observation Dataset}

So far we have presented our proposed method in a general context, also discussing the implicit possibilities for estimating feature importance and building ensembles. In order to demonstrate the effectiveness of our method, we apply it in the increasingly relevant field of Climate Studies.

Our planet is a perfect example of a multi-layered, highly interconnected system, called the Earth System~\cite{steffen2020emergence}. In this paper, we will use NASA's Earth Observation (NEO) Dataset, containing satellite measurements of various climate factors from all around the globe. For our purposes, we chose to work with 19 such layers, namely: Vegetation Index (NVDI), Snow Cover (SNOWC), Daytime Temperature (LSTD), Nighttime Temperature (LSTN), Cloud Optical Thickness (COT), Cloud Particle Radius (LCD\_RD), Cloud Fraction (CLD\_FR), Cloud Water Content (CLD\_WP), Nitrogen Dioxide (NO2), Ozone (OZONE), Chlorophyll (CHLORA), Sea Surface Temperature (SST), Fire (FIRE), Leaf Area Index (LAI), Daytime Temperature Anomaly (LSTD\_AN), Nighttime Temperature Anomaly (LSTN\_AN), Aerosol Optical Depth (AOD), Carbon Monoxide (CO), and Water Vapor (WV).

The NEO Dataset provides measurements in the form of $540\times1080$ maps. We split these maps into $288$ patches of size $45\times45$ and compute averages on these patches in order to reduce the data load. We sample measurements monthly for the time span 2004--2022, and keep the last 30 months before 2017 for testing. Each layer is normalized independently such that the mean is $0$ and the standard deviation is $1$.

\clearpage

\subsection{Observing the Distribution Shift and Implementing a Selection Algorithm}

We begin by analyzing the prediction accuracy for a MAE model trained with uniform masking percentages. The nodes we use are: AOD, CHLORA, CLD\_FR, CLD\_RD, CLD\_WP, COT, LSTD, LSTN, NO2, NVDI, OZONE, SNOWC, SST. We train the model on the years 2004--2022, but reserve the last 43 months for testing (note that, for this experiment, we change the test time span in order for the experiment to reflect most recent years). In order to observe a potential distribution shift, we compute the accuracies of each prediction in chronological order and plot them on a graph.

\begin{figure}[H]
\centering
\includegraphics[scale=0.15]{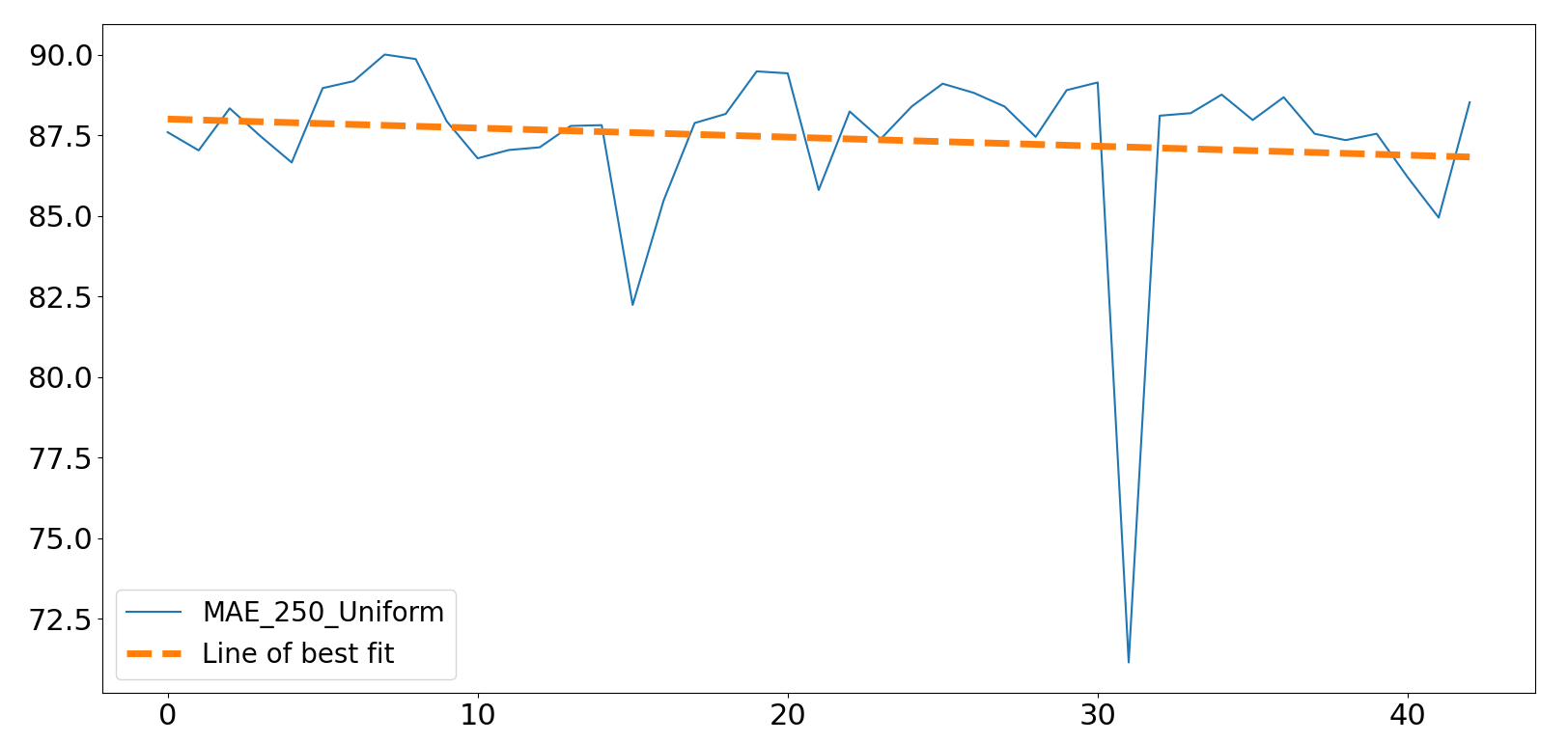}
\caption{The plot shows the prediction accuracies as we move away from the last seen training case. If there were little to no climatical changes, we would expect no significant decay in accuracy, but plotting the line of best fit on the graph clearly shows a steady decrease in accuracy, which points towards Climate Change}
\label{fig:acc_trend}
\end{figure}

By analyzing the behaviour of our MAE, we get a clear indication that there is indeed a distribution shift in the dataset. Therefore, we seek to better understand this development, and we introduce a method for analyzing local distribution shifts. Unlike the previous method which reflected the global distribution shift over all layers and locations on Earth, our next method is used on a more detailed local scale (specific layers and locations), both to determine the amplitude of the shift, and to visualize it in the form of a 2D plot as it shifts over time.

We start by creating 12-dimensional points for every year for a particular patch. The coordinates of such a point represent the monthly measurements for that year of some predetermined climate factor. Then, we can reduce the dimensionality to 2D by implementing Principal Component Analysis.

This is done for every patch on a particular layer. Moreover, we introduce a metric of variability by considering the largest Eigenvalue (corresponding to the first PCA component) for each such patch. This metric of variability is highlighted on the global map, and for any selected patch, we also observe the distribution shift in 2D.

\begin{figure}[H]
\centering
\includegraphics[scale=0.35]{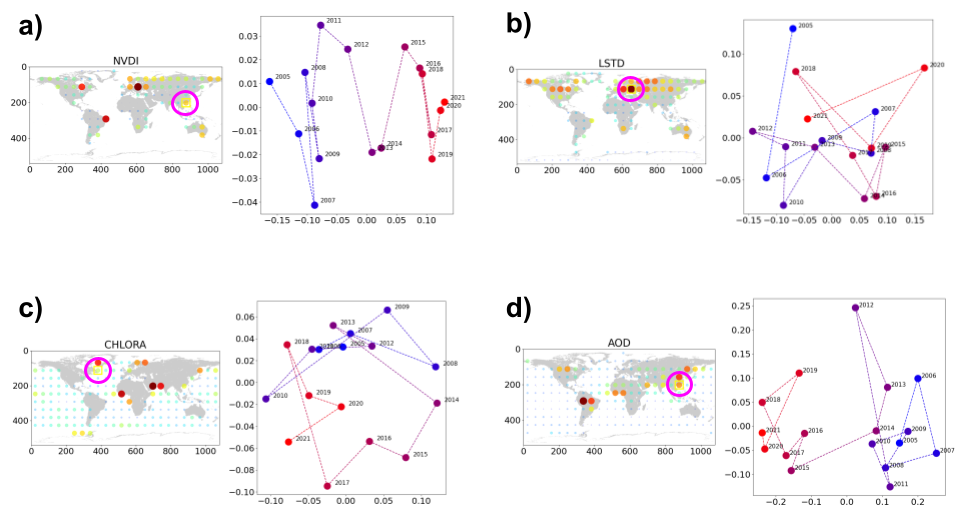}
\caption{We perform the analysis on the four different layers: NVDI ({\it a}), LSTD ({\it b}) CHLORA ({\it c}) and AOD ({\it d}). The figure shows the global map highlighted based on the first eigenvalue ({\it left}), from which we select some patches for PCA ({\it yellow box in magenta circle}). Note that high variance regions (indicating larger shifts) are mostly around highly populated areas and at the border between continent and ocean. We project the 12-dimensional data points (corresponding to the yellow patch on the left) on the first two PCA components and plot them in 2D describing the years 2005-2021 ({\it blue} to {\it red}). Note, for example, how in plot ({\it a}) the points move from left (years close to 2005) to right (years close to 2021), clearly indicating a climate distribution shift.}
\label{fig:PCA}
\end{figure}

In order to focus on the highly variable patches for our next experiments, we now implement a selection algorithm based on the maximal Eigenvalues computed above. The selection is made by eliminating all patches that are adjacent to another patch of higher variability (on edge or corner). Performing this selection algorithm on our dataset results in 349 highly variable patches that are suitable for our experiments.

Note that the selection method described in section 2.2 could have also been used at this step. However, the purpose of that feature selection method is to maintain a high prediction accuracy while eliminating large amounts of features. This is not the main focus in our experiments, so we chose to use this other method in order to focus on highly variable locations that are harder to predict, and instead, to use the method in section 2.2 in order to analyze feature importance in section 3.6.

\subsection{Creating Ensembles for Semi-supervised Learning}

Before discussing our main experiments, we also describe our procedure for enhancing performance using semi-supervised learning. Ensembles, besides improving the accuracy of our method on their own, represent a very suitable way of creating pseudo-labels for semi-supervised learning. We therefore describe a way to train any model on a semi-supervised dataset created using any other model:
\begin{enumerate}
    \item First, we train the first model on the labeled training dataset.
    \item Then, we generate labels for the unlabeled test dataset using the first model. However, we do not use the model itself, but the aggregated predictions generated by an ensemble as described in section 2.3.
    \item We append the resulting pseudo-labeled samples to the original training dataset.
    \item Finally, we train the second model on the pseudo-labeled combined dataset.
\end{enumerate}

\begin{figure}[H]
\centering
\includegraphics[scale=0.2]{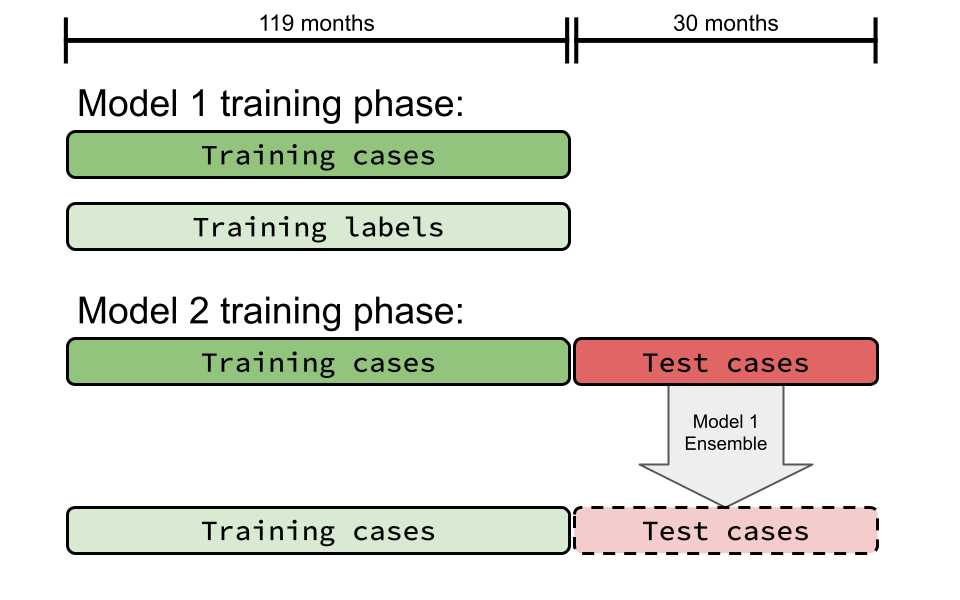}
\caption{The datasets used for standard training and semi-supervised training (ratio ignored for readability). The green bars represent the training cases ({\it dark green}) and their corresponding labels ({\it light green}), whereas the red bars represents the test cases ({\it dark red}) and the pseudo-labels generated by Model 1 ({\it light red})}
\label{fig:semi-supervised}
\end{figure}

This way, any model has the chance to benefit from a boost in performance by having access to an additional pseudo-labeled dataset to learn from.

\subsection{Observing the Efficiency of Ensemble Boosting for Different Masking Percentages}

Now we begin experimenting with different settings and make comparisons between standard models such as Multi-Layer Perceptrons, Masked Autoencoders (with and without ensemble boosting), as well as the same models, but trained through semi-supervised learning with the use of ensembles. The following layers were selected as output layers: FIRE, LAI, LSTD\_AN, LSTN\_AN, AOD, CO, WV.

The accuracy is calculated as $1$, minus the L1 loss. Therefore, $100\%$ represents a perfect match of the prediction with the ground truth, while $0\%$ represents the estimated accuracy of predicting the layer-wise average each time (because the standard deviation of data is $1$).

First, we seek to understand how different masking percentages when building ensembles correlate with the accuracy of the model. We compare two models with the same number of parameters: a simple MLP and a MAE model. However, we also add to the comparison the ensemble created on the MAE model, and we do the experiment for different masking percentages of the ensemble.

\begin{figure}[H]
\centering
\begin{minipage}{0.49\textwidth}
\includegraphics[width=1\linewidth]{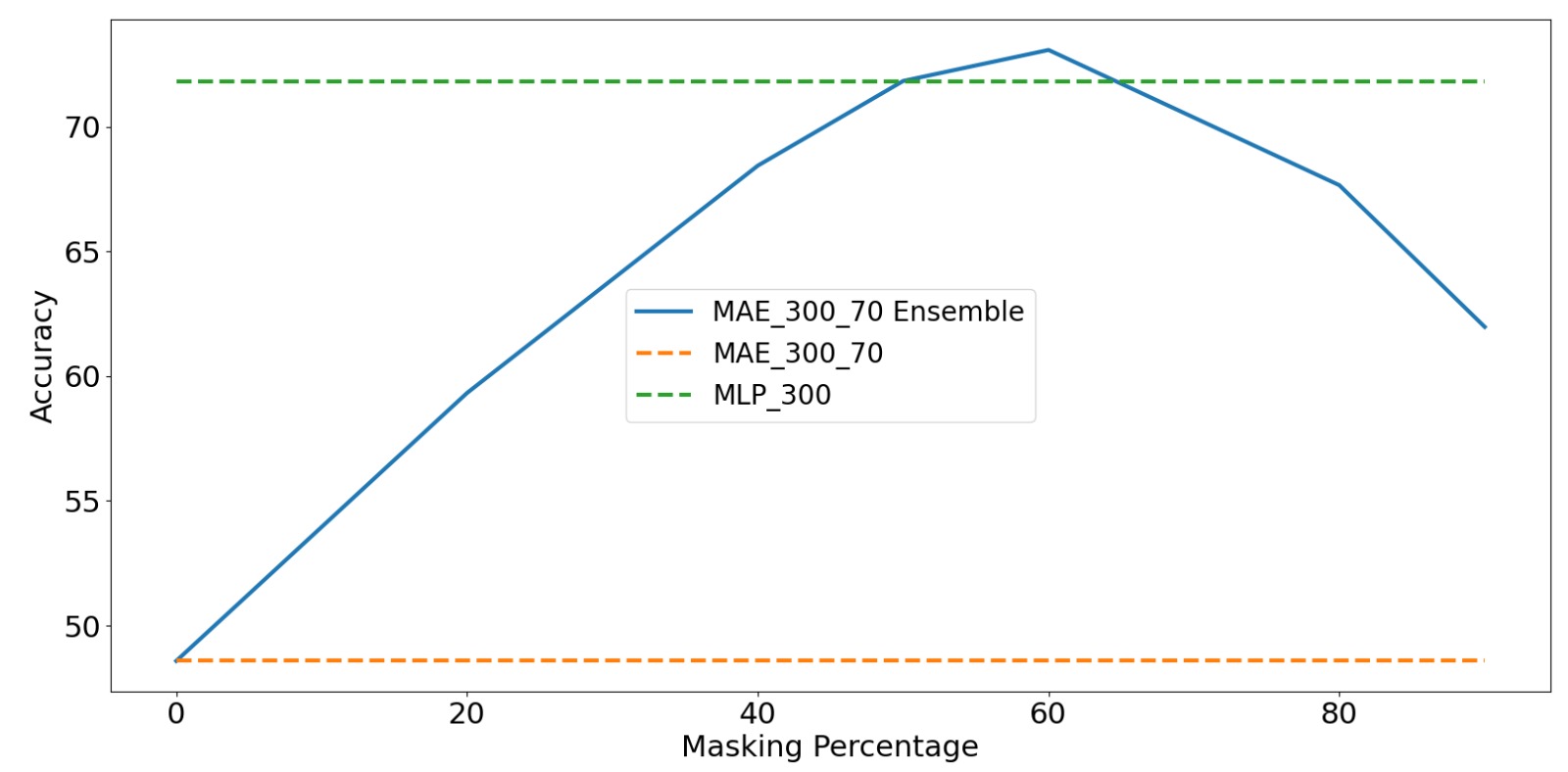}
\end{minipage}
\begin{minipage}{0.49\textwidth}
\includegraphics[width=1\linewidth]{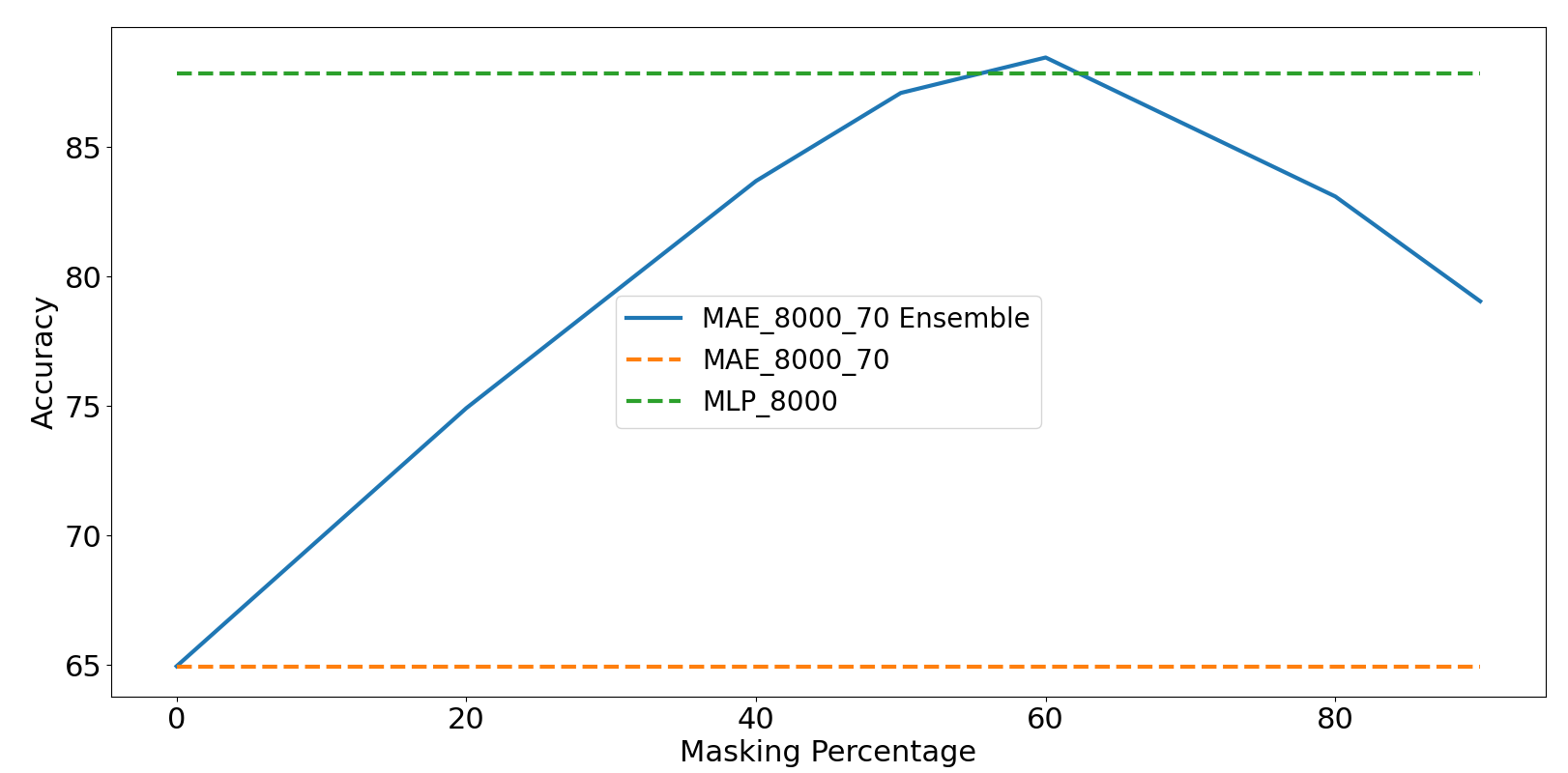}
\end{minipage}
 \caption{The graph shows the accuracy of two different models: an MLP and a MAE ({\it dotted lines}), both on the selected patches ({\it left}) and the full image ({\it right}). It is expected that the task-specific MLP outperforms the MAE on an input-output scenario, but when we apply an ensemble to the MAE ({\it blue line}), we observe a massive boost in performance, even surpassing the MLP in the case of $60\%$ masking}\label{Fig:acc_ensemble}
\end{figure}

The graphs (both for the high variability dataset and the full image dataset) clearly show the advantages of using the ensemble built over the MAE:
\begin{enumerate}
    \item The multiple random masking ensemble method results in a great increase in accuracy, leveraging on the highly adaptive nature of the MAE.
    \item The resulting ensemble became more powerful than a standard MLP. Unlike the MLP, the ensemble, and the MAE itself, are not in danger of over-fitting, as their sample pool is in effect exponential in the number of features.
    \item The MAE is trained on the general case of reconstructing all features when $70\%$ random maskings are applied, while the MLP is completely task-specific. If we were to change the input-output split of the test, we would therefore experience similar accuracy without the need to re-train.
\end{enumerate}

\clearpage

\subsection{Semi-supervised Learning}

In this experiment we boost different models using our semi-supervised learning approach described in section 3.2. The model used for creating pseudo-labels is the ensemble with $60\%$ masking built over another MAE trained with $70\%$ masking. In this case, the MAE could just as well be trained with uniform masking, but since semi-supervised learning implies task-awareness when generating pseudo-labels, training the MAE with a specific $70\%$ masking percentage can allow for an extra boost in label quality. Note that the MAE models trained on the new dataset are still task-blind.

In our comparison we introduce the following models. First, a simple MAE trained with uniform masking, alongside the ensemble built over it. We also include the ensemble built over a MAE trained with 70\% masking. These 3 models are task-blind. We also introduce three task-specific standard models: a simple MLP with the same number of parameters as the MAE, Linear Regression, as well as the Lasso Regressor.

\setlength{\tabcolsep}{4pt}
\begin{table}
\begin{center}
\caption{Accuracy comparison for the 6 models presented above. For each of them, the accuracy is calculated when the model was trained on the labeled dataset, and also when the model was trained on the semi-supervised dataset. Observe the boost in performance for almost all models, especially the massive increase in performance for Linear Regression.}
\label{table:accuracy}
\begin{tabular}{llcc}
\hline
\noalign{\smallskip}
\multicolumn{2}{c}{\textbf{Model}} & \textbf{Supervised} & \textbf{Semi-supervised}\\
\noalign{\smallskip}
\hline
\noalign{\smallskip}
\multirow{3}{*}{Task-blind} & MAE Uniform & 68.92 & \textbf{69.91} \\
& MAE Uniform Ensemble & 70.53 & \textbf{72.79} \\
& MAE 70\% Mask Ensemble & \textbf{73.08} & 72.86 \\
\noalign{\smallskip}
\hline
\noalign{\smallskip}
\multirow{3}{*}{Task-specific} & MLP & 71.84 & \textbf{73.42} \\
& Linear Regression & 54.22 & \textbf{73.10} \\
& Lasso Regressor & 73.09 & \textbf{74.07} \\
\noalign{\smallskip}
\hline
\end{tabular}
\end{center}
\end{table}
\setlength{\tabcolsep}{1.4pt}

\subsection{Comparing Prediction Accuracy while Gradually Masking Data}

What makes our approach much more powerful and attractive, is the capacity to adapt to any input-output relation, without having to know in advance which layers will be input and which will be output. Essentially, we would be able, under our general strategy, to reconstruct any layer or any part of it, from any other available data, no matter how much or how little we provide.

In order to verify this claim, we compare five models: Lasso Regressor, AdaBoost, a simple MLP and two MAE models, the second of which was boosted through semi-supervised learning. The MLP and the MAEs have the same number of parameters. To test the robustness of each method, we randomly mask a percentage of features, and we perform the experiment while gradually increasing the percentage from $0\%$ to $95\%$.

\begin{figure}[H]
\centering
\includegraphics[scale=0.15]{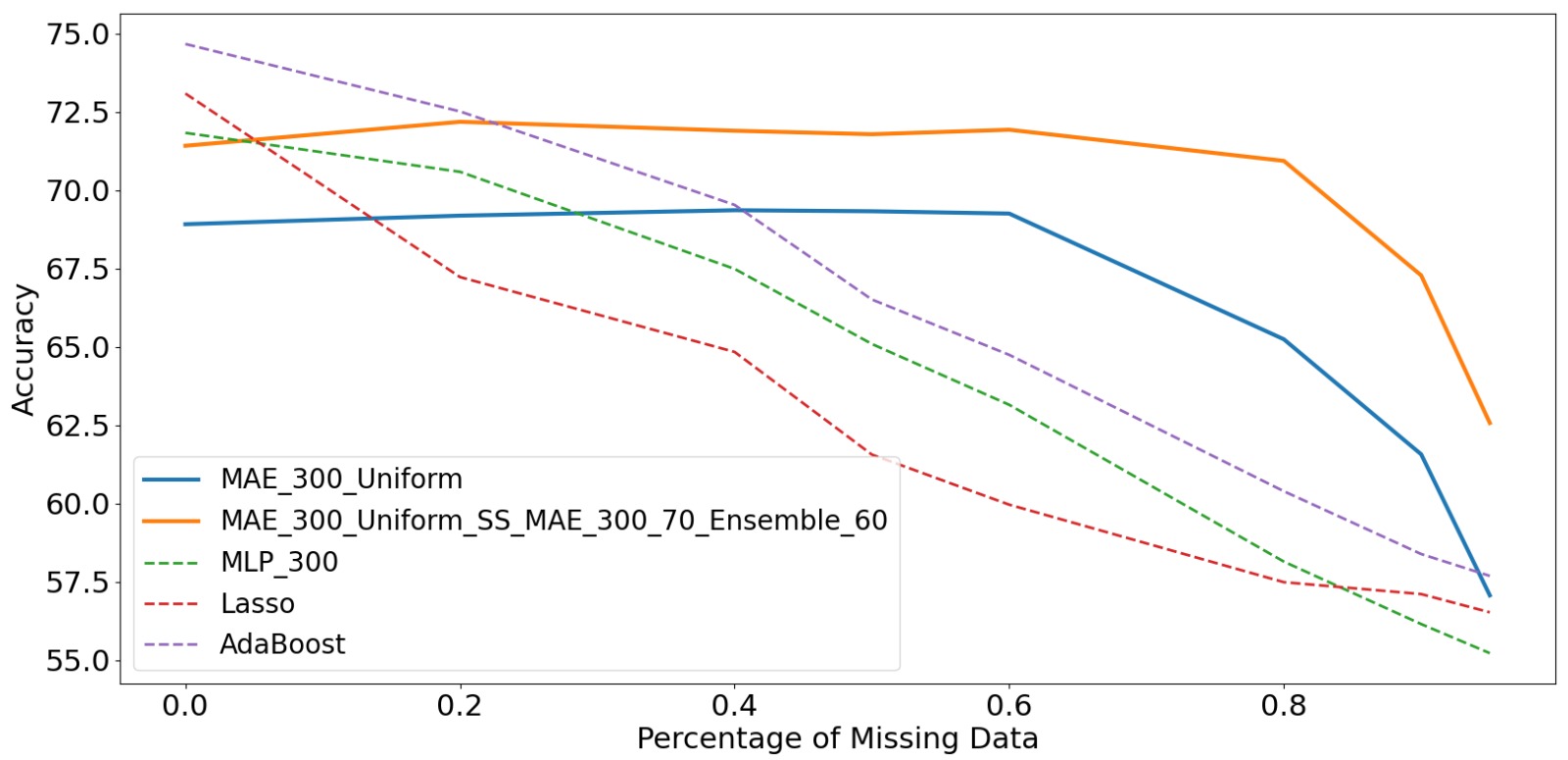}
\caption{The graph shows the accuracy of five different models as the amount of missing data is increasing. Observe the decline in accuracy of standard methods such as MLP, Lasso and AdaBoost, while the MAE models remain constant until 60\% masking. Moreover, semi-supervised learning allowed the second MAE to remain virtually constant until 80\% masking}
\label{fig:acc_decay}
\end{figure}

\begin{figure}[H]
\centering
\includegraphics[scale=0.15]{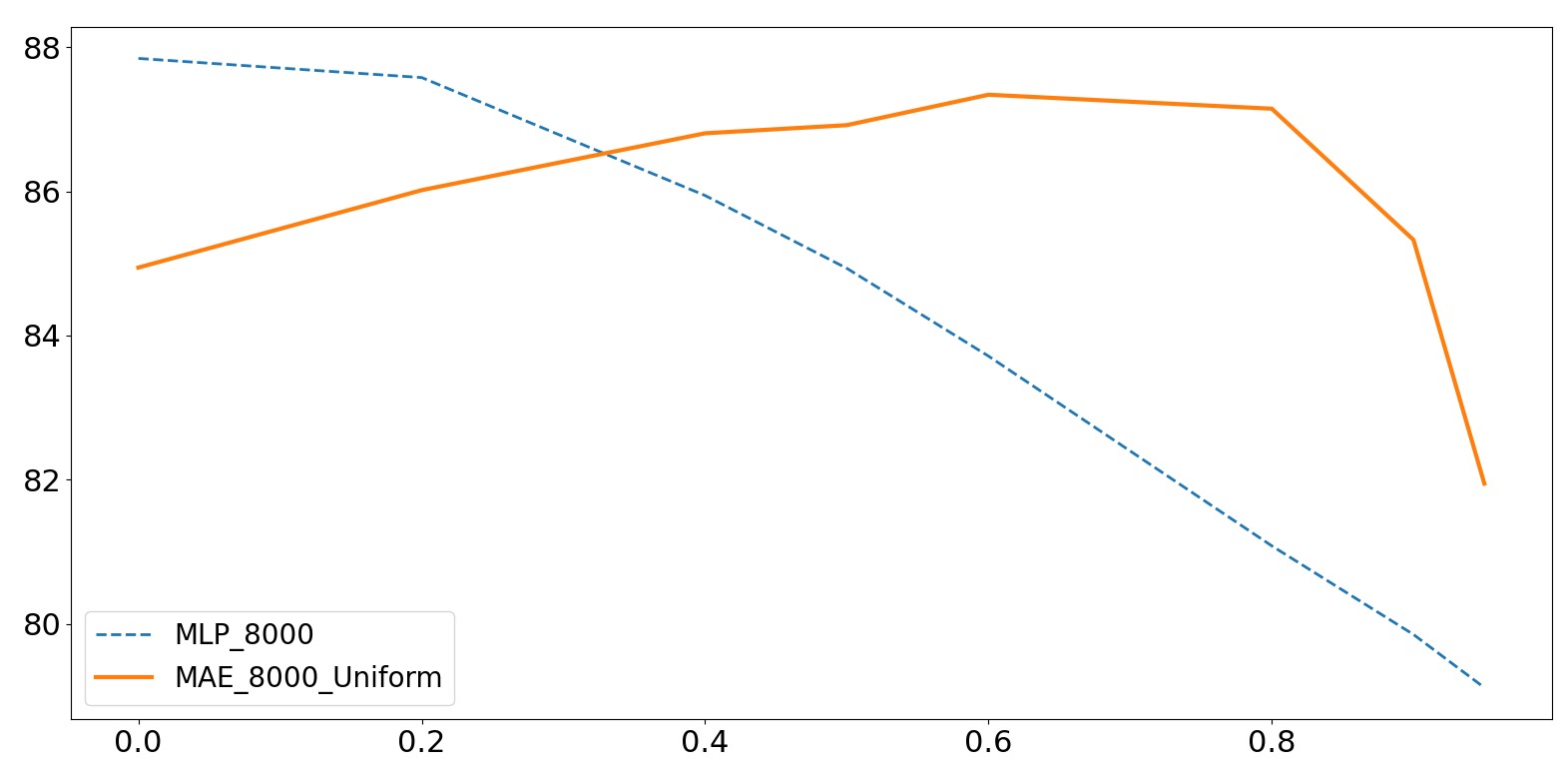}
\caption{The graph represents the equivalent experiment performed on the full dataset. We focused on only two models: an MLP and a MAE trained with uniform masking. Note that, in this experiment, not only does the MAE outperform the MLP on larger masking percentages, but it peaks in performance at a $60\%$ missing data, probably due to its training. Therefore, the same accuracy can obviously be achieved when less data is missing, for example, by artificially masking the remaining percentage to $60\%$}
\label{fig:acc_decay_full}
\end{figure}

\clearpage

\subsection{Feature Importance Estimation and Interpretation}

Next, we test our learning approach for the task of measuring feature importance and present some interesting insights -- results that strongly suggest that our approach could be very useful for Climate Research. We estimate feature importance by computing the \textbf{Loss Matrix} as described earlier in the paper.

For this experiment, we leave the patch selection aside and move to the scenario of making predictions on the full map.

\begin{figure}[H]
\centering
\includegraphics[scale=0.25]{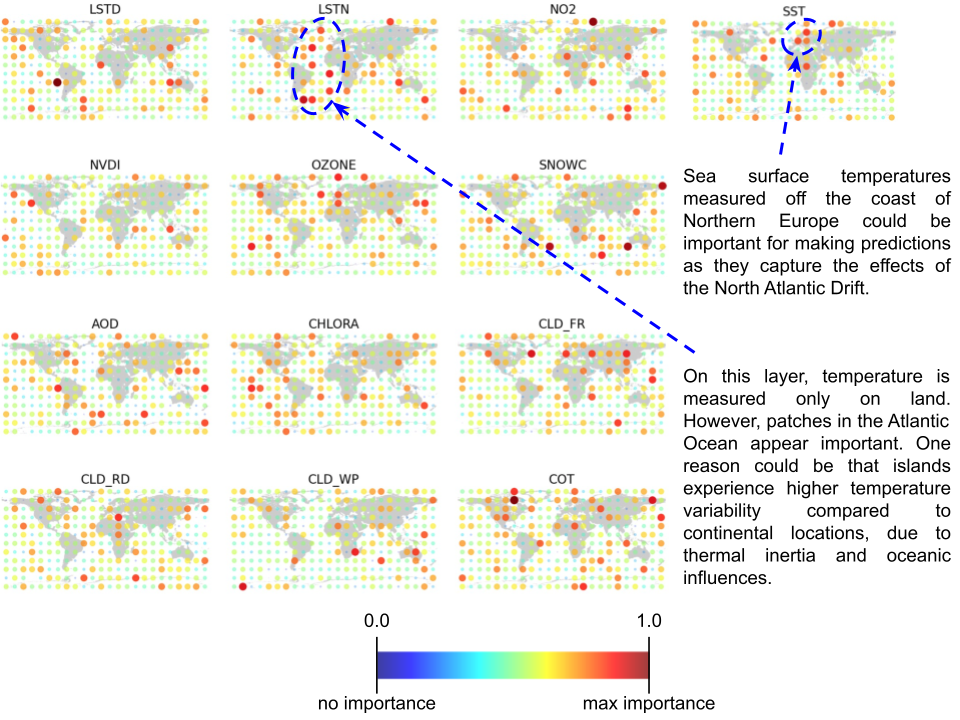}
\caption{The general importance of each feature is highlighted on the map. Observe the two regions of highly important patches and their respective possible explanations}
\label{fig:importance}
\end{figure}

We compute the \textbf{Global} average loss as described earlier, and multiply the value by $-1$ to get a metric of importance instead (lower average loss when the feature is present translates directly to higher importance). Then, we normalize the values to span the range $[0,1]$, and plot the results on the map for each layer.

In order to interpret the feature importance metric correctly, we make the following claim. The importance of a specific location and layer represents its ability to capture meaningful and important climatic processes that are \textbf{unique} to a small number of such patches. As a counter-example, the general seasonal movements are experienced in almost all locations and climate layers, so the presence of a specific unmasked patch in the input pool will not drastically decrease loss by gaining unique insights into this behaviour. In contrast, our method thrives in uncovering small-scale, local behaviours that have an effect on the Earth's Climate. 

\clearpage

\subsection{Comparison With the Multi-Task Hyper-graph Model}

Finally, we compare our approach with the more powerful Multi-Task Hyper-graph Model~\cite{pirvu2023multi}, which is trained for the more difficult task of predicting low-scale measurements in the Earth System, whereas our model is trained to predict average measurements on patches of size $45\times45$. In order to provide a fair comparison we translate the predictions generated by teh Hyper-graph Model to the higher scale scenario by computing patch averages on their predictions. Then, we compute the accuracy metric for both methods on the test dataset, for each of the nodes FIRE, LAI, LSTD\_AN, LSTN\_AN, AOD, CO and WV.

\begin{figure}[H]
\centering
\includegraphics[scale=0.3]{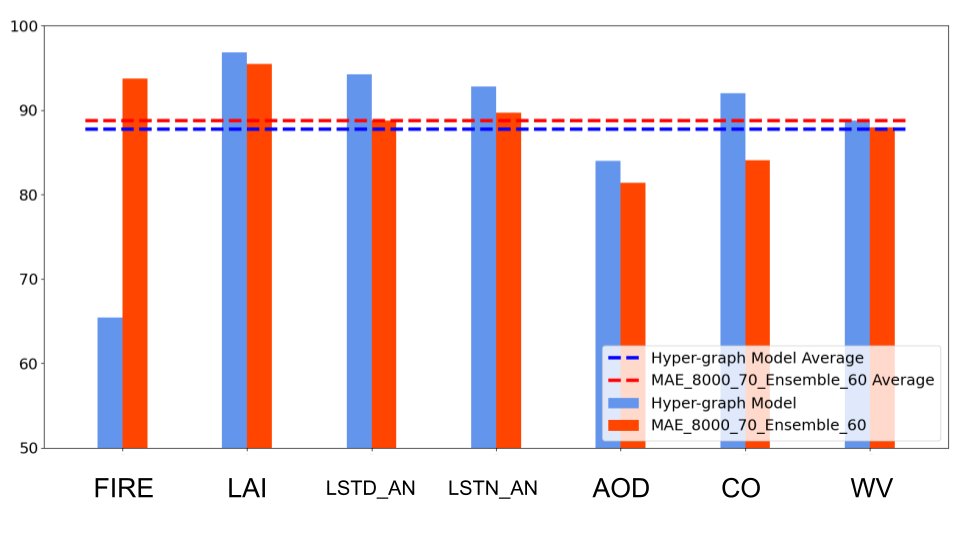}
\caption{The figure shows the accuracies achieved by the two models on each output node, as well as the averages over them. It is expected that the more complex model used by them outperforms our MLP-based model. Still, our method is not far behind, and even achieves a higher accuracy on the FIRE node, which represents a highly difficult climate factor to predict on a low scale. Thus, our method is robust, and performs better on average}
\label{fig:compare}
\end{figure}

The results are encouraging. Even when compared to a more complex model, trained on high-resolution images, our method performs well, which leads us naturally to our next step -- replacing the simple MLP used as the base model in our approach with a more powerful model. 

\section{Discussion and Conclusions}

We presented MR--MAE, a novel ensemble multi-modal learning method based on masked auto-encoders with multiple random masking, such that each random masking case becomes a candidate in a large pool of MAEs that form the final ensemble. Each MAE is, by training, robust to missing data and implicitly learns a mapping between any combination of input (non-masked data) and output (the masked data). 

Our multimodal approach is able to learn correlations among many different multiple layers (modalities) and is perfectly suited for the task of Earth Observation and prediction from satellite data, where we can have tens of different types of modalities.
Moreover, our approach is also suitable for semi-supervised learning, since the ensemble can produce reliable pseudo-labels on which we can distill a single MAE model. Our experiments demonstrate the effectiveness of semi-supervised learning for Earth Observation by using as pseudo-labels the output produced by the Multiple Random Masking ensembles.

Additionally, we also show how the Multiple Random Masking approach can be effectively used for computing the relevance of each input feature (token or local patch from a certain observation channel) in the prediction of any other feature. In all our experiments on Earth Observation data from NASA (NEO dataset) we demonstrated the capability of our approach to deal with large amounts of missing data and greatly outperform classical methods (such as classic MLPs, Lasso and Linear Regression), which, unlike our approach, were specifically trained for mapping certain input layers to the output ones.

The key strength of our approach is learning to predict across many different interpretation layers and spatial locations any set of missing parts from any set of visible ones. The number of layers we use at training time is larger than other works in the literature. By masking entire layers randomly we insure that at test time we can predict any set of output layers from any set of input ones, which is also different from existing literature. Our training approach also gives the possibility to form robust ensembles at no additional training cost. From this perspective our method is related to recent Test-time Augmentation (TTA) algorithms~\cite{prakash2018deflecting,ashukha2020pitfalls,zhang2022memo}, but, to our best knowledge, no other TTA method used masking to form ensembles.

Overall, our proposed method is simple, effective and successfully addresses challenges from several dimensions in AI, such as learning with missing multi-modal data, ensemble learning and semi-supervised learning, feature importance estimation and selection. From a practical perspective, our model is also well suited for better understanding the Earth System, by learning to predict any missing climate observations and to discover teleconnection influences across climate factors and between distant locations, as a complementary way to study climate, from a more global perspective. 

While our specific implementation was based on MLPs, which we chose for ease of training and testing, our approach is general, therefore testing much more powerful Transformer models on large quantities of data will be our next step.

\vspace{-0.25cm}

\section{Acknowledgements} 
\vspace{-0.25cm}
This work was supported in part by the EU Horizon project ELIAS (No. 101120237). We want to express our sincere gratitude towards
Mihai Pirvu (University Politehnica of Bucharest) for his help with the NEO Dataset and to Elena Burceanu (Bitdefender) and Iulia Popescu (University of Oxford) for the helpful technical discussions.
\clearpage

\bibliographystyle{splncs}
\bibliography{egbib}

\begin{thebibliography}{10}

\bibitem{he2022masked}
He, K., Chen, X., Xie, S., Li, Y., Doll{\'a}r, P., Girshick, R.:
\newblock Masked autoencoders are scalable vision learners.
\newblock In: Proceedings of the IEEE/CVF conference on computer vision and pattern recognition. (2022)  16000--16009

\bibitem{mizrahi20234m}
Mizrahi, D., Bachmann, R., Kar, O.F., Yeo, T., Gao, M., Dehghan, A., Zamir, A.:
\newblock 4m: Massively multimodal masked modeling.
\newblock arXiv preprint arXiv:2312.06647 (2023)

\bibitem{bachmann2022multimae}
Bachmann, R., Mizrahi, D., Atanov, A., Zamir, A.:
\newblock Multimae: Multi-modal multi-task masked autoencoders.
\newblock In: European Conference on Computer Vision, Springer (2022)  348--367

\bibitem{chen2019self}
Chen, Y., Schmid, C., Sminchisescu, C.:
\newblock Self-supervised learning with geometric constraints in monocular video: Connecting flow, depth, and camera.
\newblock In: Proceedings of the IEEE International Conference on Computer Vision. (2019)  7063--7072

\bibitem{zhou2017unsupervised}
Zhou, T., Brown, M., Snavely, N., Lowe, D.G.:
\newblock Unsupervised learning of depth and ego-motion from video.
\newblock In: Proceedings of the IEEE Conference on Computer Vision and Pattern Recognition. (2017)  1851--1858

\bibitem{ranjan2019competitive}
Ranjan, A., Jampani, V., Balles, L., Kim, K., Sun, D., Wulff, J., Black, M.J.:
\newblock Competitive collaboration: Joint unsupervised learning of depth, camera motion, optical flow and motion segmentation.
\newblock In: Proceedings of the IEEE Conference on Computer Vision and Pattern Recognition. (2019)  12240--12249

\bibitem{bian2019unsupervised}
Bian, J., Li, Z., Wang, N., Zhan, H., Shen, C., Cheng, M.M., Reid, I.:
\newblock Unsupervised scale-consistent depth and ego-motion learning from monocular video.
\newblock In: Advances in Neural Information Processing Systems. (2019)  35--45

\bibitem{gordon2019depth}
Gordon, A., Li, H., Jonschkowski, R., Angelova, A.:
\newblock Depth from videos in the wild: Unsupervised monocular depth learning from unknown cameras.
\newblock In: Proceedings of the IEEE/CVF International Conference on Computer Vision. (2019)  8977--8986

\bibitem{yang2018unsupervised}
Yang, Z., Wang, P., Xu, W., Zhao, L., Nevatia, R.:
\newblock Unsupervised learning of geometry from videos with edge-aware depth-normal consistency.
\newblock In: Thirty-Second AAAI Conference on Artificial Intelligence. (2018)

\bibitem{tosi2020distilled}
Tosi, F., Aleotti, F., Ramirez, P.Z., Poggi, M., Salti, S., Stefano, L.D., Mattoccia, S.:
\newblock Distilled semantics for comprehensive scene understanding from videos.
\newblock In: Proceedings of the IEEE/CVF Conference on Computer Vision and Pattern Recognition. (2020)  4654--4665

\bibitem{guizilini2020semantically}
Guizilini, V., Hou, R., Li, J., Ambrus, R., Gaidon, A.:
\newblock Semantically-guided representation learning for self-supervised monocular depth.
\newblock arXiv preprint arXiv:2002.12319 (2020)

\bibitem{chen2019towards}
Chen, P.Y., Liu, A.H., Liu, Y.C., Wang, Y.C.F.:
\newblock Towards scene understanding: Unsupervised monocular depth estimation with semantic-aware representation.
\newblock In: Proceedings of the IEEE Conference on Computer Vision and Pattern Recognition. (2019)  2624--2632

\bibitem{stekovic2020casting}
Stekovic, S., Fraundorfer, F., Lepetit, V.:
\newblock Casting geometric constraints in semantic segmentation as semi-supervised learning.
\newblock In: The IEEE Winter Conference on Applications of Computer Vision. (2020)  1854--1863

\bibitem{Hu_2019_CVPR}
Hu, D., Nie, F., Li, X.:
\newblock Deep multimodal clustering for unsupervised audiovisual learning.
\newblock In: The IEEE Conference on Computer Vision and Pattern Recognition (CVPR). (June 2019)

\bibitem{Li_2019_CVPR}
Li, Y., Zhu, J.Y., Tedrake, R., Torralba, A.:
\newblock Connecting touch and vision via cross-modal prediction.
\newblock In: The IEEE Conference on Computer Vision and Pattern Recognition (CVPR). (June 2019)

\bibitem{zhang2017split}
Zhang, R., Isola, P., Efros, A.A.:
\newblock Split-brain autoencoders: Unsupervised learning by cross-channel prediction.
\newblock In: Proceedings of the IEEE Conference on Computer Vision and Pattern Recognition. (2017)  1058--1067

\bibitem{pan2004automatic}
Pan, J.Y., Yang, H.J., Faloutsos, C., Duygulu, P.:
\newblock Automatic multimedia cross-modal correlation discovery.
\newblock In: Proceedings of the tenth ACM SIGKDD international conference on Knowledge discovery and data mining, ACM (2004)  653--658

\bibitem{he2017unsupervised}
He, L., Xu, X., Lu, H., Yang, Y., Shen, F., Shen, H.T.:
\newblock Unsupervised cross-modal retrieval through adversarial learning.
\newblock In: 2017 IEEE International Conference on Multimedia and Expo (ICME), IEEE (2017)  1153--1158

\bibitem{zhao2018sound}
Zhao, H., Gan, C., Rouditchenko, A., Vondrick, C., McDermott, J., Torralba, A.:
\newblock The sound of pixels.
\newblock In: Proceedings of the European Conference on Computer Vision (ECCV). (2018)  570--586

\bibitem{zamir2018taskonomy}
Zamir, A.R., Sax, A., Shen, W., Guibas, L.J., Malik, J., Savarese, S.:
\newblock Taskonomy: Disentangling task transfer learning.
\newblock In: Proceedings of the IEEE conference on computer vision and pattern recognition. (2018)  3712--3722

\bibitem{zamir2020robust}
Zamir, A.R., Sax, A., Cheerla, N., Suri, R., Cao, Z., Malik, J., Guibas, L.J.:
\newblock Robust learning through cross-task consistency.
\newblock In: Proceedings of the IEEE/CVF Conference on Computer Vision and Pattern Recognition. (2020)  11197--11206

\bibitem{fifty2021efficiently}
Fifty, C., Amid, E., Zhao, Z., Yu, T., Anil, R., Finn, C.:
\newblock Efficiently identifying task groupings for multi-task learning.
\newblock Advances in Neural Information Processing Systems \textbf{34} (2021)  27503--27516

\bibitem{marcu2023self}
Marcu, A., Pirvu, M., Costea, D., Haller, E., Slusanschi, E., Belbachir, A.N., Sukthankar, R., Leordeanu, M.:
\newblock Self-supervised hypergraphs for learning multiple world interpretations.
\newblock In: Proceedings of the IEEE/CVF International Conference on Computer Vision. (2023)  983--992

\bibitem{pirvu2023multi}
Pirvu, M., Marcu, A., Dobrescu, M.A., Belbachir, A.N., Leordeanu, M.:
\newblock Multi-task hypergraphs for semi-supervised learning using earth observations.
\newblock In: Proceedings of the IEEE/CVF International Conference on Computer Vision. (2023)  3404--3414

\bibitem{leordeanu2021semi}
Leordeanu, M., P{\^\i}rvu, M.C., Costea, D., Marcu, A.E., Slusanschi, E., Sukthankar, R.:
\newblock Semi-supervised learning for multi-task scene understanding by neural graph consensus.
\newblock In: Proceedings of the AAAI Conference on Artificial Intelligence. Volume~35. (2021)  1882--1892

\bibitem{steffen2020emergence}
Steffen, W., Richardson, K., Rockstr{\"o}m, J., Schellnhuber, H.J., Dube, O.P., Dutreuil, S., Lenton, T.M., Lubchenco, J.:
\newblock The emergence and evolution of earth system science.
\newblock Nature Reviews Earth \& Environment \textbf{1}(1) (2020)  54--63

\bibitem{prakash2018deflecting}
Prakash, A., Moran, N., Garber, S., DiLillo, A., Storer, J.:
\newblock Deflecting adversarial attacks with pixel deflection.
\newblock In: Proceedings of the IEEE conference on computer vision and pattern recognition. (2018)  8571--8580

\bibitem{ashukha2020pitfalls}
Ashukha, A., Lyzhov, A., Molchanov, D., Vetrov, D.:
\newblock Pitfalls of in-domain uncertainty estimation and ensembling in deep learning.
\newblock arXiv preprint arXiv:2002.06470 (2020)

\bibitem{zhang2022memo}
Zhang, M., Levine, S., Finn, C.:
\newblock Memo: Test time robustness via adaptation and augmentation.
\newblock Advances in Neural Information Processing Systems \textbf{35} (2022)  38629--38642

\end{thebibliography}
\end{document}